\documentclass[letterpaper,10pt,conference]{ieeeconf}
\usepackage{cw}
\usepackage{tikz}
\IEEEoverridecommandlockouts
\graphicspath{{images/},{diagrams/}}
\begin{acronym}[MVSEC] 
\acro{HCI}{Human-Computer Interaction}
\acro{ML}{Machine Learning} 
\acro{SLAM}{Simultaneous Localization And Mapping} 
\acro{TTI}{Time-To-Impact}
\acro{IoU}{Intersection over Union}
\acro{MVSEC}{Multivehicle Stereo Event Camera}
\acro{VIO}{visual-inertial odometry}
\acro{AEE}{average endpoint error}
\acro{EE}{endpoint error}
\end{acronym}

\title{\LARGE\bf{}EVReflex: Dense Time-to-Impact Prediction\\for Event-based Obstacle Avoidance}

\author{Celyn Walters\(^1\) and Simon Hadfield\(^1\)
\thanks{*
	This work was supported by the UK Engineering and Physical Sciences Research Council (EPSRC) grant agreement EP/S035761/1 ``Reflexive robotics''.
	We would also like to thank Dr.~Hamideh Kerdegari for her help and initial work on this topic.
}%
\thanks{\(^1\)Celyn Walters and Simon Hadfield are with CVSSP, University of Surrey, UK
{\tt\small celyn.walters@surrey.ac.uk}}%
}
\newcommand\copyrighttext{%
	\footnotesize \textcopyright 2021 IEEE. Personal use of this material is permitted.
	Permission from IEEE must be obtained for all other uses, in any current or future
	media, including reprinting/republishing this material for advertising or promotional
	purposes, creating new collective works, for resale or redistribution to servers or
	lists, or reuse of any copyrighted component of this work in other works.
}
\newcommand\copyrightnotice{%
	\begin{tikzpicture}[remember picture,overlay]
	\node[anchor=south,yshift=10pt] at (current page.south) {\fbox{\parbox{\dimexpr\textwidth-\fboxsep-\fboxrule\relax}{\copyrighttext}}};
	\end{tikzpicture}%
}
\begin{document}
\maketitle
\copyrightnotice
\thispagestyle{empty}
\pagestyle{empty}

\begin{abstract}\label{abstract}
The broad scope of obstacle avoidance has led to many kinds of computer vision-based approaches.
Despite its popularity, it is not a solved problem.
Traditional computer vision techniques using cameras and depth sensors often focus on static scenes, or rely on priors about the obstacles.
Recent developments in bio-inspired sensors present event cameras as a compelling choice for dynamic scenes.
Although these sensors have many advantages over their frame-based counterparts, such as high dynamic range and temporal resolution, event-based perception has largely remained in 2D.
This often leads to solutions reliant on heuristics and specific to a particular task.

We show that the fusion of events and depth overcomes the failure cases of each individual modality when performing obstacle avoidance.
Our proposed approach unifies event camera and lidar streams to estimate metric \ac{TTI} without prior knowledge of the scene geometry or obstacles.
In addition, we release an extensive event-based dataset with six visual streams spanning over 700 scanned scenes.

\end{abstract}

\section{INTRODUCTION}\label{sec:introduction}

Obstacle avoidance is a cornerstone of autonomous robotic behaviour.
It enables robots to explore unknown and dynamic environments safely, as well as being vital for \ac{HCI}.
As such, it has been a topic of interest in the field of computer vision since its inception.
Vision-based techniques have historically relied on salient cues such as image discontinuities and corner features.
This causes vision-based approaches to perform less well in poor lighting or where there is little or repetitive texture.
In these scenarios, many motion planning implementations utilise depth sensors like lidar, which mitigate issues with texture, lighting and motion blur.
However, limited resolution and slow scan times make it challenging to evade small or fast moving objects.
This is particularly an issue in domains like autonomous driving, where fast reaction times are vital.

Event cameras are an emerging technology which are a promising alternative or supplement to traditional frame-based cameras and depth sensors.
Rather than producing entire image frames, event cameras output a stream of per-pixel intensity changes.
This has much higher temporal resolution and dynamic range than regular RGB images, and almost eliminates motion blur.
The bandwidth of the event stream can widen or narrow depending on the amount of change observed.
These properties are clearly useful for the task of obstacle avoidance, but due to the different data composition, specialised techniques must be created to utilise these cameras.

\begin{figure}[t]\centering
	\subcaptionbox{
		Depth sensors (high-latency)
		\label{fig:front:depth}
	}[\linewidth]{
		\includegraphics[width=0.49\linewidth]{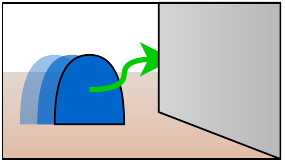}
		\hfill
		\includegraphics[width=0.49\linewidth]{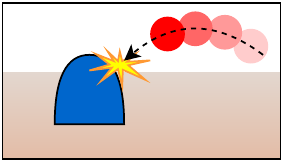}
	}
	\\
	\subcaptionbox{
		Event cameras (low-latency)
		\label{fig:front:events}
	}[\linewidth]{
		\includegraphics[width=0.49\linewidth]{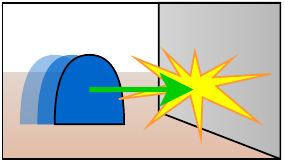}
		\hfill
		\includegraphics[width=0.49\linewidth]{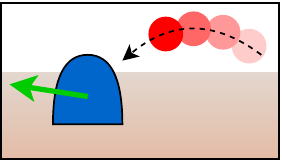}
	}
	\\
	\subcaptionbox{
		Fused depth and events (ours)
		\label{fig:front:both}
	}[\linewidth]{
		\includegraphics[width=0.49\linewidth]{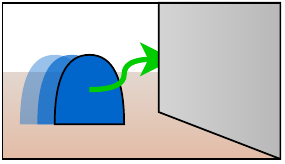}
		\hfill
		\includegraphics[width=0.49\linewidth]{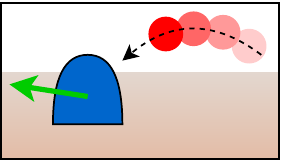}
	}
	\caption{Illustration of sensor capabilities for collision avoidance.
	Depth cameras work well for large static objects but struggle to detect small fast-moving objects.
	Event cameras can segment dynamic foreground objects but cannot infer depth to avoid large textureless obstacles.
	Our work demonstrates their fusion works in both cases.}
	\label{fig:front}
	\vspace{-1\baselineskip}
\end{figure}

Similar to traditional frame-based cameras, event cameras operate on a 2D image plane and require dedicated structure from motion~\cite{Kim2016} or \ac{ML} techniques~\cite{Hidalgo-Carrio2020,Zhu2019} to infer depth, introducing delays.
The lack of direct depth estimation is a shortcoming for obstacle avoidance, as it is often not possible to calculate the time or location of potential collisions, and as a result, the danger to be associated to given objects.
Additionally, as with frame based cameras, it is impossible for an event camera to observe flat coloured textureless objects like walls.
The lack of intensity changes causes no events to be generated in these regions.

Similar to human stereo vision, the complementary properties of events and depth have the potential to enable robust obstacle avoidance across a previously infeasible range of scenarios.
Much like a biological creature, we argue that it is also important for the agent to be able to prioritize either depth perception or intensity changes depending on the situation.
As exemplified by \Cref{fig:front}, this enables an agent to avoid obstacles with low texture and also those moving quickly, which would be challenging for a single sensor modality.
To this end, we make the following contributions:
\begin{enumerate}
	\item An extensive dataset for developing event-based computer vision techniques\footnote{
		\label{fn:dataset}
		The dataset, code, and instructions for running and generating additional data are available at \href{https://gitlab.eps.surrey.ac.uk/cw0071/EVReflex}{https://gitlab.eps.surrey.ac.uk/cw0071/EVReflex}
	}:
	This includes RGB images, events, as well as ground truth depth, optical flow, semantic masks, and dense inverse \ac{TTI} maps.
	The extracted data covers over 700 scanned scenes from the ScanNet dataset~\cite{Dai2017} with the inclusion of additional flying obstacles.
	The data are formatted both as image databases and as ROS messages.
	\item To our knowledge, the first approach to unify lidar and event camera sensor streams within a deep-learning framework, as shown in \Cref{fig:front},
	\item Demonstration that the above approach can be used to develop systems which avoid a broader range of dangerous obstacles than any previous technique.
\end{enumerate}

\section{RELATED WORK}\label{sec:literature}

\noindent\textbf{Obstacle avoidance} ---
Fundamentally, solving obstacle avoidance in static scenes comprises of pose estimation and mapping.
As such, robotic obstacle avoidance has shared roots with visual odometry and \ac{SLAM}.
The differences between them are often only in the framing of the problem.
Early approaches such as the work of \citeauthor{Borenstein1989} utilise a certainty grid and potential field for collision-free local path planning~\cite{Borenstein1989}.
More recently, approaches such as the work of \citeauthor{Chakravarty2017} use depth estimation from single images~\cite{Chakravarty2017}.
These perform reasonably well in clear open environments but tend to over-smooth small and distinct details.
The majority of approaches only consider static scenes, in which RGB and depth sensors are suitable.
We propose that the fusion of depth with an event stream is more suitable where there is a mixture of static and dynamic obstacles.

\noindent\textbf{Events} ---
While the event representation has inherent advantages, it must be processed differently from traditional images.
Event-based approaches for many classic computer vision problems have emerged recently.
\Citeauthor{Zhu2018} estimate optical flow from events, using a self-supervised network supervised by corresponding greyscale images~\cite{Zhu2018}.
In \citeauthor{Hidalgo-Carrio2020}'s work, a monocular event stream is used in a recurrent network to estimate depth using temporal consistency.
EVDodge from \citeauthor{Sanket2019} is an approach which uses events to estimate `segmentation flow', optical flow with foreground/background segmentation~\cite{Sanket2019}.
They train in simulation and show a drone capable of dodging thrown objects.

\noindent\textbf{Event camera datasets} ---
Following the recent availability of event cameras, a few datasets have emerged each with a slightly different focus.
For \ac{VIO}, \Citeauthor{Mueggler2017} introduced an event-based dataset which includes images and IMU, with ground truth poses~\cite{Mueggler2017}.
The popular \ac{MVSEC} dataset from \Citeauthor{Zhu2018a} followed, which involves stereo event cameras and additionally provides lidar point clouds and more scene variety~\cite{Zhu2018a}.
These have enabled successful approaches \cite{Zhu2018,Zhu2019,Hidalgo-Carrio2020}, but they lack extra data streams like ground truth optical flow and semantic labels.

Simulated datasets, while having a domain gap to be overcome for real-world applications, can provide accurate ground truth to supervise or evaluate other approaches.
Early simulators emulated events by processing differences between image frames~\cite{Kaiser2017}.
\Citeauthor{Mueggler2017} used a custom rendering engine for accurate event data in simple scenes~\cite{Mueggler2017}.
InteriorNet from \citeauthor{Li2018} boasts many additional streams such as optical flow and surface normals, in realistic customisable scenes~\cite{Li2018}.
However, \cite{Mueggler2017,Li2018} still operate on frames.
\Citeauthor{Rebecq2018} enhanced \cite{Mueggler2017} with ESIM, which uses adaptive-rate sampling~\cite{Rebecq2018} but does not match the realism and flexibility of InteriorNet.
We use real scanned scenes from the ScanNet dataset~\cite{Dai2017} in conjunction with ESIM for a varied dataset which correctly models an event camera.

\section{METHODOLOGY}\label{sec:methodology}

\begin{figure}[t]\centering
	\includegraphics[width=\linewidth]{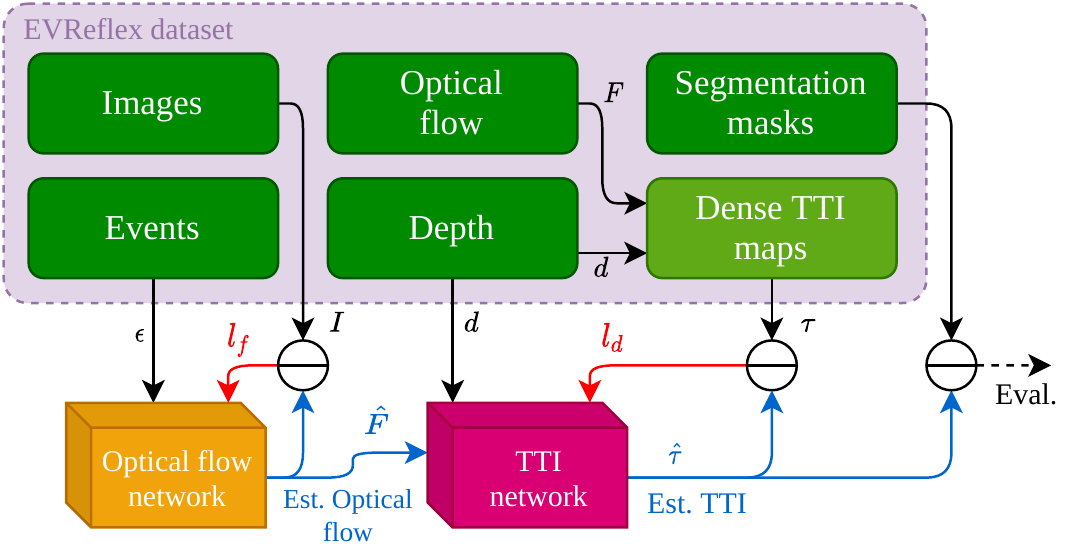}
	\caption{
		Dataset and network data flow.
		The \(\ominus\) represents the application of the relevant loss function.
		Loss signals are notated in red, output signals are notated in blue.
	}
	\label{fig:structure}
	\vspace{-1\baselineskip}
\end{figure}
As shown in \Cref{fig:structure}, our approach consists of two parts.
Firstly, the event stream \(\epsilon\) is used to predict optical flow \(\hat{F}\) between image frames.
The optical flow network \(O\) is self-supervised using the greyscale image stream \(I\) which corresponds with the events.
In the second part, the resulting optical flow is combined with depth \(d\) and processed by the second subnetwork \(P\).
The \ac{TTI} estimation network directly estimates inverse metric \ac{TTI}, the supervisory signal being formed from depth frames warped by optical flow.

\subsection{Optical flow estimation}\label{sec:methodology:flow}

\begin{figure}[t]\centering
	\includegraphics[width=\linewidth]{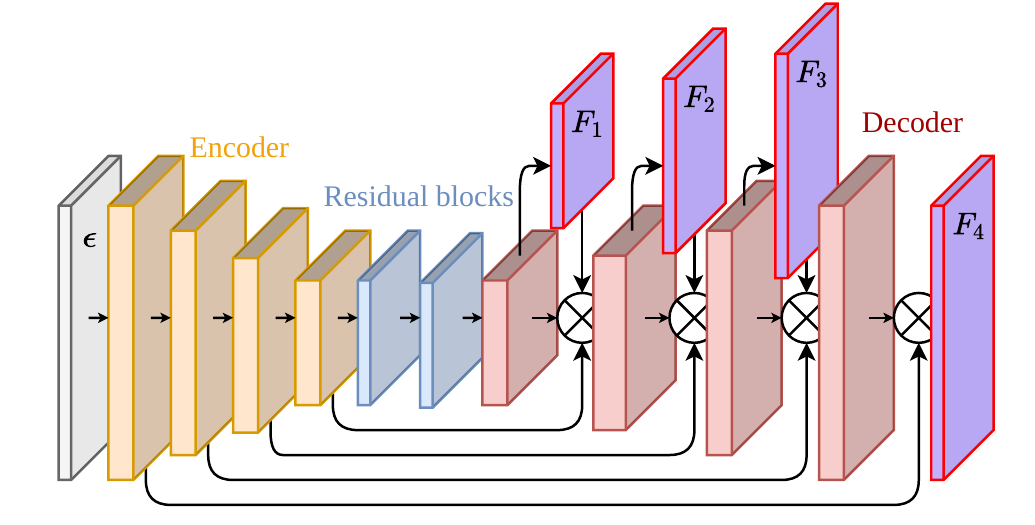}
	\caption{
		Flow estimation network architecture.
		The \(\bigotimes\) represents a concatenation operation.
		Total loss is calculated as the mean of losses from each decoder layer's output \(F_{1-4}\).
	}
	\label{fig:flownet}
	\vspace{-1\baselineskip}
\end{figure}

The task of optical flow estimation has had a long history of approaches based on traditional visible images.
Inherent drawbacks with the characteristics of this data, such as motion blur, fixed framerates and poor low light performance, affect all of these approaches.
While many hand-crafted techniques mitigate these problems~\cite{Weinzaepfel2013,Revaud2015}, event cameras naturally perceive small changes with high temporal resolution.
The performance of event cameras for optical flow estimation has been shown to exceed traditional cameras particularly for thin structures and low-texture areas~\cite{Zhu2018}.
These advantages are particularly important for obstacle avoidance, where we can expect fast-moving objects to exhibit motion blur in traditional images.

Inspired by the work of \citeauthor{Zhu2018}~\cite{Zhu2018}, the first module of our approach is a network which estimates the optical flow for use in our second module (\Cref{sec:methodology:tti}).
This network is trained in a self-supervised manner, using the grayscale images in the training loss function.
This means that ground truth optical flow is not required to train for a new dataset.

Although events themselves are streamed asynchronously between pixels, many works opt to treat them similar to image frames.
This is often implemented as an accumulation of events between \(t\) and \(t - 1\).
It is known that both the positive/negative event counts and event times are informative signals for event-based learning~\cite{Nguyen2019}.
Following the work of \citeauthor{Zhu2018}, we represent the accumulated events \(\epsilon\) as a 4 channel image \([4, H, W]\).
The channels of this event map are the positive and negative event count at that pixel (\(\epsilon^{n+}\), \(\epsilon^{n-}\)) and the timestamps of the most recent positive/negative event (\(\epsilon^{t+}\), \(\epsilon^{t-}\)).
This compact format has been shown to be sufficient for many event-based image processing tasks~\cite{Lagorce2017,Park2016}.
Resembling a U-Net~\cite{Ronneberger2015}, our network \(O\) is composed of four encoder and four decoder layers with skip connections, shown in \Cref{fig:flownet}.
Given network weights \(\theta_O\), the result is an estimated flow field \(F\) at each decoder layer:
\begin{equation}
	F = O(\epsilon | \theta_O)
	.
\end{equation}
The application of \(F\) on the corresponding grayscale image \(I_t\) is compared with the next image \(I_{t + 1}\) using photometric and smoothness losses.
The photometric loss \(l_p\) minimises intensity differences with Charbonnier loss~\cite{Sun2014} \(\rho\):
\begin{equation}
	\label{eq:photometric}
	\begin{array}{ccc}
		l_p(F, I_t, I_{t + 1}) = \sum\limits_i{\rho\left(I_t(i) - I_{t + 1}(\mathcal{W}(i | F))\right)} & \forall i \in I
	\end{array}
	,
\end{equation}
where \(i\) is a pixel coordinate \(\left\{x, y\right\}\) and \(\mathcal{W}\) warps an image by sampling it according to a flow field.
\(\rho\) is calculated using \(\epsilon{=}0.001, \alpha{=}0.45\).
The smoothness loss \(l_s\) regularises flow discontinuities in a pixel's neighbourhood \(N\),
\begin{equation}
	\label{eq:smooth}
	\begin{array}{ccc}
		l_s(F) = \sum\limits_i \sum\limits_{n \in N}\rho\left(i - n\right) & \forall i \in F
	\end{array}
	.
\end{equation}
The two losses are combined as a weighted sum, \(l_f = l_p + \alpha l_s\), with \(\alpha\) as the weight.
By substituting \Cref{eq:photometric,eq:smooth}, the training process for the self-supervised flow network in isolation is
\begin{equation}
	\label{eq:flowloss}
	\hat{\theta_O} = \argmin_{\theta_O}\left(l_p(O(\epsilon | \theta_P), I_t, I_{t + 1}) + \alpha l_s(O(\epsilon | \theta_O))\right)
	.
\end{equation}

\subsection{\ac{TTI} estimation}\label{sec:methodology:tti}

\begin{figure}[t]\centering
	\includegraphics[width=\linewidth]{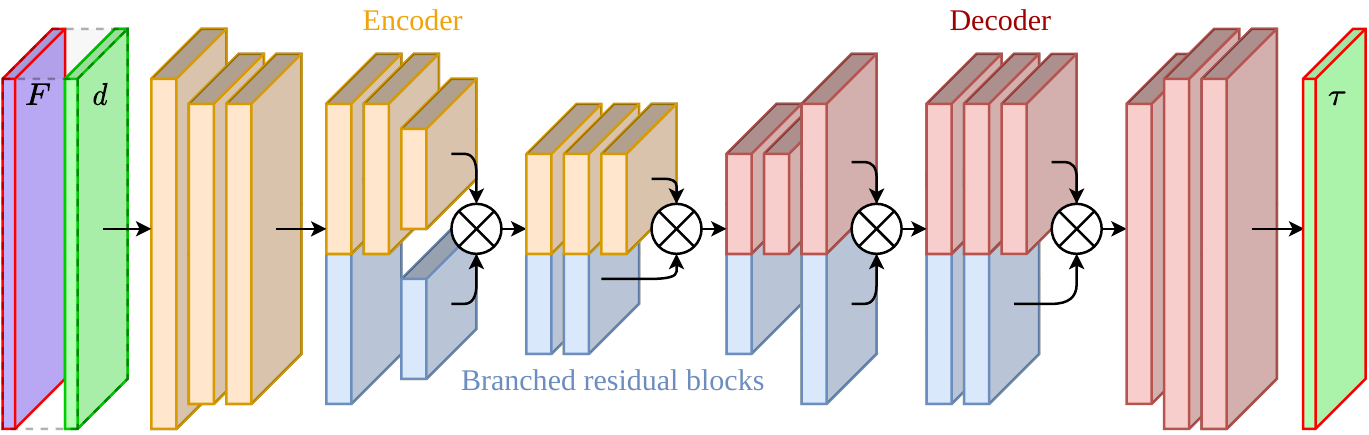}
	\caption{
		Inverse \ac{TTI} estimation network architecture.
	}
	\label{fig:ttinet}
	\vspace{-1\baselineskip}
\end{figure}

The task of this network is to regress the inverse \ac{TTI} \(\hat{\tau}\) for each pixel given optical flow \(F\) and depth values \(d\).
Inverse \ac{TTI} is used to avoid the asymptote in \ac{TTI} at the point where the relative egocentric velocity between the agent and an obstacle is zero.
To investigate the network's reliance between optical flow and instantaneous depth, we define two versions, and each are tested independently in \Cref{sec:experiments}.
The first, which we refer to as the `static' version, takes the stacked optical flow \(F_t\) and instantaneous depth \(d_t\) as input.
Here, the speed of geometry moving into the image plane must be estimated implicitly during the computation of the TTI map.
The second version additionally uses the next sequential depth frame \(d_{t + 1}\) to provide additional cues about motion in the Z axis.
We refer to this version as the `dynamic' version.

The network \(P\), shown in \Cref{fig:ttinet}, downsamples the input twice through 9 convolutional layers with branched residual connections, and performs the transposed version to upsample back to the original resolution.
The network is supervised by the ground truth inverse \ac{TTI} \(\tau\), explained in \Cref{sec:methodology:dataset}, using mean-squared error loss \(l_\tau\) such that
\begin{equation}
	\label{eq:ttiloss}
	\begin{array}{>{\(}W{c}{0.45\linewidth}<{\)}>{\(}W{c}{0.45\linewidth}<{\)}}
		\hat{\theta_P} = \argmin\limits_{\theta_P}\left(l_\tau(\hat{\tau}, \tau)\right) & \where \hat{\tau} = P(\left[F, d\right] | \theta_P)
		.
	\end{array}
\end{equation}

\subsection{Control policy}\label{sec:methodology:control}

In the work of \citeauthor{Sanket2019}, some obstacle detection is based on priors.
For completely unknown objects, the input to the control policy is the optical flow after masking segmented background regions~\cite{Sanket2019}.
The obstacle trajectory is assumed to be the average direction of the masked optical flow on the image plane, and evasion is directed to be perpendicular to this.
There is no check that this motion is safe; the priority is to avoid a fast-moving object.

For the comparison in \Cref{sec:experiments:evasion}, we take a similar approach without any priors on the obstacles.
We combine the depth with predicted \ac{TTI} into a 3D motion vector, and then use the cross-product of this with the agent's egomotion \(v\) as the direction of evasion \(\psi\),
\begin{equation}
	\label{eq:control}
	\psi = \left(\frac{1}{n} \sum\limits^n_{i=1}\left(F(i), d(i)\hat{\tau}(i)\right)\right) \times v
	.
\end{equation}
Because our network makes a prediction for every pixel, we can additionally avoid colliding with `background' regions if they are within the field of view.

\subsection{Dataset}\label{sec:methodology:dataset}

A dataset, EVReflex, was created to facilitate evaluation of our approach.
With its extensive ground truth for multiple sensors, it is also a useful asset for future event camera-based research.

To collect this dataset, the movement of a robotic agent is simulated in indoor scenes.
The robot is equipped with an RGB camera, a depth camera, IMU, and an event camera.
The simulator also outputs the pose of the agent and ground truth optical flow~\cite{Rebecq2018}, and has been extended to be capable of producing semantic class labels.
At the time of writing, there are very few other datasets which provide an event stream as well as semantic labels.

A problem for many synthetic datasets is the domain gap between simulated environments and the real world, due to the lack of realism.
This issue applies to most event camera datasets including EVDodge~\cite{Sanket2019} and those based on ESIM~\cite{Rebecq2018}.
EVReflex overcomes this major problem by simulating in RGB-D scanned scenes from the ScanNet dataset~\cite{Dai2017}, with a total of over 700 different real-world rooms.
The simulated robot follows random floor-based trajectories\footnote{
	Most translational variation is in the \([X, Y]\) floor plane, and most rotational variation in \(Z\) rotation (yaw).
} within each room, which brings it into proximity with various static obstacles.
Concurrent with this, a few dynamic obstacles are simulated, each of which follow randomized trajectories that periodically bring them into proximity with the robot.

A notable contribution of this dataset to aid future research on collision avoidance tasks is the additional image stream for the identification of dangerous image regions.
Each pixel \(i\) in these images represents inverse \acf{TTI} values for that point in the scene.
These are computed by comparing the previous depth \(d_{t - 1}\) warped by the optical flow \(F\) with the current depth \(d_t\),
\begin{equation}
	\label{eq:tti}
	\tau(i) = \max\left(0, \frac{d_t(i) - d_{t - 1}(\mathcal{W}(i | F))}{d_t}\right)
	.
\end{equation}
This is fundamentally different from foreground/background segmentation such as the estimated in EVDodge~\cite{Sanket2019}.
Firstly, this approach can function with an arbitrary number of obstacles of any shape.
There is no intrinsic preference between compact dynamic obstacles and static geometry.
Secondly, objects which are not approaching the robot, including those in close proximity, have zero values.

The dataset is available to download\cref{fn:dataset} in both image databases and ROS formats.
We also provide the full simulation system to allow people to customise and generate additional data if needed.

\section{EXPERIMENTS}\label{sec:experiments}

The evaluation is split into three parts.
Firstly, evaluation was performed on the performance of the optical flow network.
Secondly the accuracy of the dense \ac{TTI} regression.
This evaluation was broken down across different semantic classes to highlight the performance for different types of obstacles.
Where applicable, we compare against the state-of-the-art in event-based obstacle avoidance (EVDodge~\cite{Sanket2019}), and against a depth-sensor baseline, which thresholds nearby objects as obstacles.
Finally, we evaluate the accuracy of the signal for the control policy across the baselines.

The EVReflex dataset was split into training, evaluation and test splits at a ratio of 70\%, 15\% and 15\% respectively.
This split was performed across the environments, meaning that the same environment is not visible in more than one set.

The optical flow estimation subnetwork (\Cref{sec:methodology:flow}) was trained with a initial learning rate of \num{1e-5}, with a batch size of \num{8} for \num{30} epochs.

The \ac{TTI} estimation subnetwork (\Cref{sec:methodology:tti}) was trained for \num{40} epochs, with a batch size of \num{40} and a learning rate of \num{0.01} which reduced by a factor of \num{10} on a validation loss plateau.

\subsection{Event-based optical flow evaluation}\label{sec:experiments:flow}

\begin{figure}[t]\centering
	\includegraphics[width=0.25\columnwidth]{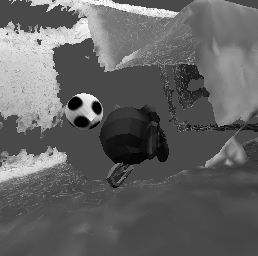}%
	\includegraphics[width=0.25\columnwidth]{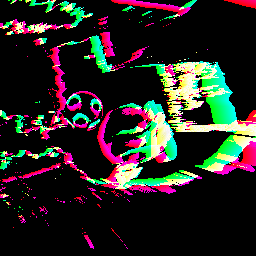}%
	\includegraphics[width=0.25\columnwidth]{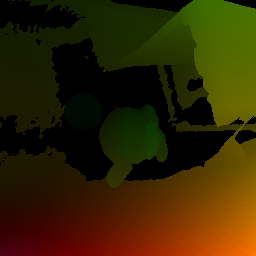}%
	\includegraphics[width=0.25\columnwidth]{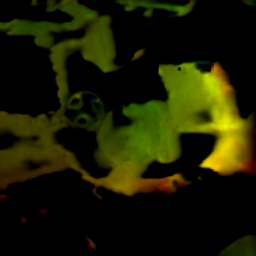}%
	\\
	\includegraphics[width=0.25\columnwidth]{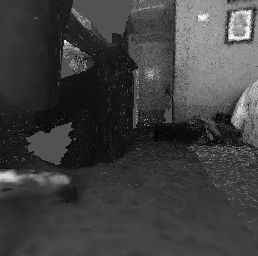}%
	\includegraphics[width=0.25\columnwidth]{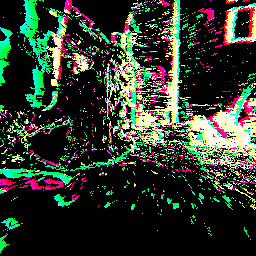}%
	\includegraphics[width=0.25\columnwidth]{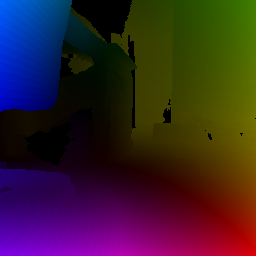}%
	\includegraphics[width=0.25\columnwidth]{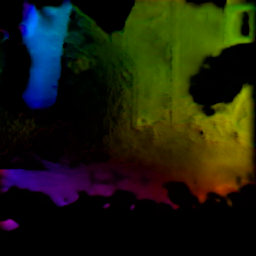}%
	\\
	\subcaptionbox{\centering{}Greyscale image}[0.25\linewidth]{\includegraphics[width=\linewidth]{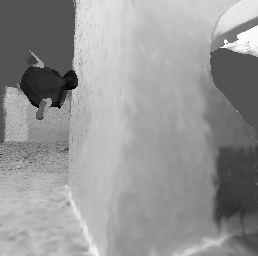}}%
	\subcaptionbox{\centering{}Events\hspace{\textwidth}(\(+\)green, \(-\)red)}[0.25\linewidth]{\includegraphics[width=\linewidth]{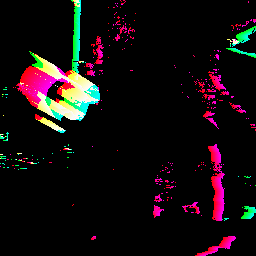}}%
	\subcaptionbox{\centering{}Ground truth optical flow}[0.25\linewidth]{\includegraphics[width=\linewidth]{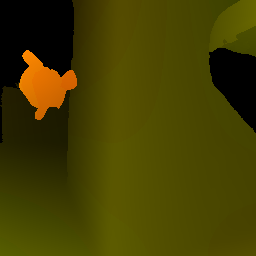}}%
	\subcaptionbox{\centering{}Estimated optical flow}[0.25\linewidth]{\includegraphics[width=\linewidth]{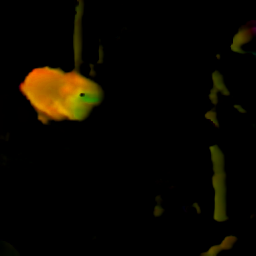}}%
	\caption{Qualitative optical flow estimation results.
	Clearly, optical flow cannot be estimated for low texture regions which do not generate events.}
	\label{fig:opticalflow}
	\vspace{-1\baselineskip}
\end{figure}

We evaluate optical flow by computing average \ac{EE} with respect to the ground truth optical flow.
By filtering to event locations only, the system achieved an overall average \ac{EE} of \num{0.36}, with \num{0.47}\% of those being outliers with an \ac{EE} of greater than 3 pixels.
However, without filtering, the average \ac{EE} is \num{15.69} with \num{84.44}\% outliers.
Our dataset utilises laser-scanned scenes from the ScanNet dataset.
The scanning process has high spatial accuracy, but textures are smoothed.
This leads to decreased events, and as a result, higher difficulty estimating optical flow.
This is especially visible in \Cref{fig:opticalflow}.
Note that regions with events are still generally correct.

This reliance on salient texture is a drawback of visual sensors in general, and is one of the motivations of our overall approach.
Depth sensors perceive textureless objects well, and \Cref{sec:experiments:tti} shows that fusing instanataneous depth with this optical flow is enough determine potential collisions.

\subsection{Dense \ac{TTI} evaluation}\label{sec:experiments:tti}

\begin{figure}[t]\centering
	\includegraphics[width=0.2\columnwidth]{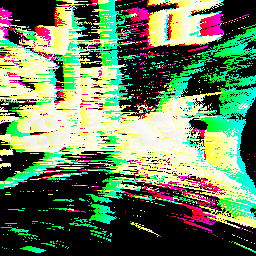}%
	\includegraphics[width=0.2\columnwidth]{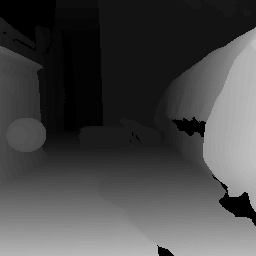}%
	\includegraphics[width=0.2\columnwidth]{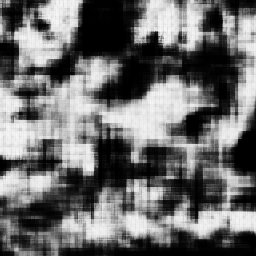}%
	\includegraphics[width=0.2\columnwidth]{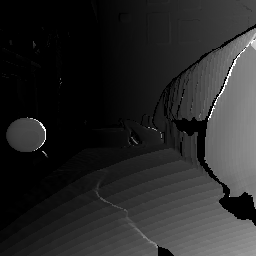}%
	\includegraphics[width=0.2\columnwidth]{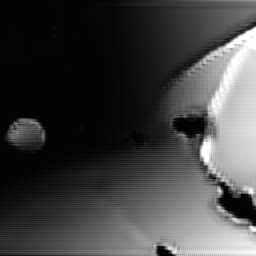}%
	\\
	\includegraphics[width=0.2\columnwidth]{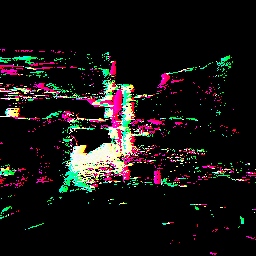}%
	\includegraphics[width=0.2\columnwidth]{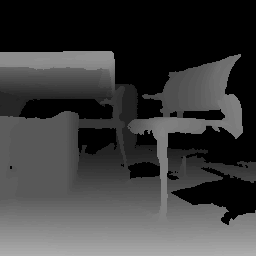}%
	\includegraphics[width=0.2\columnwidth]{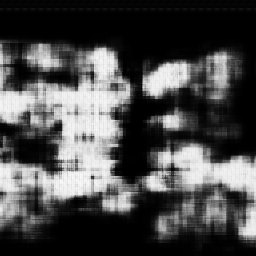}%
	\includegraphics[width=0.2\columnwidth]{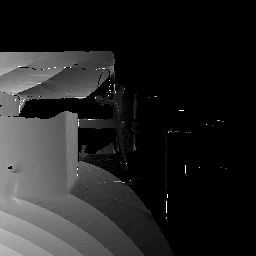}%
	\includegraphics[width=0.2\columnwidth]{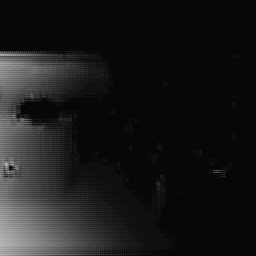}%
	\\
	\subcaptionbox{\centering{}Events}[0.2\linewidth]{\includegraphics[width=\linewidth]{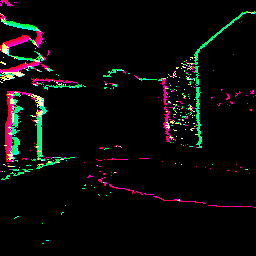}}%
	\subcaptionbox{\centering{}Depth}[0.2\linewidth]{\includegraphics[width=\linewidth]{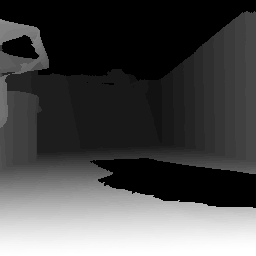}}%
	\subcaptionbox{\centering{}EVDodge predicted mask}[0.2\linewidth]{\includegraphics[width=\linewidth]{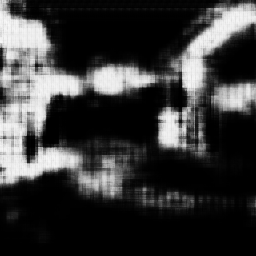}}%
	\subcaptionbox{\centering{}Inverse TTI (Ground truth)}[0.2\linewidth]{\includegraphics[width=\linewidth]{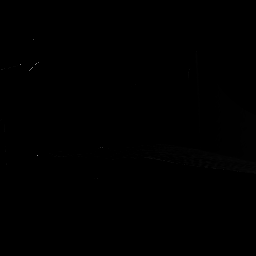}}%
	\subcaptionbox{\centering{}Inverse TTI (Estimated)}[0.2\linewidth]{\includegraphics[width=\linewidth]{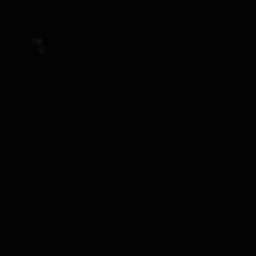}}%
	\caption{Qualitative \ac{TTI} estimation results.
	\textbf{Row 1}: Camera is moving forwards, ball is approaching camera.
	\textbf{Row 2}: Camera is rotating clockwise.
	\textbf{Row 3}: Camera is moving backwards, \ac{TTI} is negative.}
	\label{fig:results}
	\vspace{-1\baselineskip}
\end{figure}

A naïve approach to obstacle avoidance may be to consider the closest objects as the most dangerous.
This may be true for linear forward motion, but does not necessarily hold true for others.
We show a qualitative comparison between this depth baseline, the obstacle segmentation of EVDodge~\cite{Sanket2019}, and the estimated \ac{TTI} against the ground truth in \Cref{fig:results}.
EVDodge expects a clear foreground/background distinction and without it the performance is reduced.
Where the camera is rotating clockwise, static geometry on the left side of the image draws nearer while that on the right side moves further away.
Similarly, where the camera is moving backwards, \ac{TTI} does not correlate with depth.
For these cases, instantaneous depth is insufficient for collision avoidance.
The output of our approach resembles the ground truth \ac{TTI} rather than depth, and it has learned to associate motion derived from events with the depth data.
Note that we do not use explicit pose information.

To enable quantitative comparison with previous techniques based on binary segmentation of dangerous objects, we threshold our inverse \ac{TTI} values to pixels which are projected to collide within \SI{1.0}{\second}.
We compare both versions of our network against the naïve thresholded depth baseline using a value of \SI{0.5}{\metre}.
This value was chosen to give similar output to the other approaches.
We also compare against EVDodge, current state-of-the-art for event-based obstacle avoidance~\cite{Sanket2019}, for which we similarly apply a binary threshold to its segmentation output.
\Cref{tab:accuracy} shows that our approach more consistently identifies the dangerous regions.
The dataset, and as such, the task, is not suited to EVDodge, which expects clearer foreground/background delineation.
The `dynamic' version of our approach achieves very high precision but poor recall, likely as an over-reliance on depth cues.
The regular `static' version outperforms the baselines by a wide margin for recall and \(f_1\) scores.
The performance of all approaches is better for flying objects, which produce many events, than on the floor class, which produces few events.
\begin{table}[h]
	\centering
	\footnotesize
	\setlength\tabcolsep{3pt}
	\begin{tabular}{lccccccccc}
		\toprule
		\multicolumn{1}{c}{Approach} & \multicolumn{3}{c}{Flying objects} & \multicolumn{3}{c}{Floor} & \multicolumn{3}{c}{Overall}\\
		\cmidrule(lr){2-4}\cmidrule(lr){5-7}\cmidrule(lr){8-10}
		& \(p\) & \(r\) & \(f_1\) & \(p\) & \(r\) & \(f_1\) & \(p\) & \(r\) & \(f_1\)\\
		\midrule
		Depth baseline				& 0.88 & \textit{0.82} & \textit{0.85} & 0.69 & 0.26 & 0.38 & 0.55 & \textit{0.27} & \textit{0.36}\\
		EVDodge						& 0.35 & 0.40 & 0.37 & \textit{0.87} & \textit{0.27} & \textit{0.42} & \textit{0.75} & 0.22 & 0.34\\
		Ours						& \textit{0.94} & \textbf{0.89} & \textbf{0.91} & 0.38 & \textbf{0.95} & \textbf{0.54} & 0.61 & \textbf{0.95} & \textbf{0.74}\\
		Ours (dynamic)				& \textbf{1.00} & 0.59 & 0.74 & \textbf{0.99} & 0.25 & 0.41 & \textbf{0.99} & 0.22 & \textit{0.36}\\
		\bottomrule
	\end{tabular}
	\caption{Precision (\(p\)), recall (\(r\)) and F1 (\(f_1\)) scores for thresholded \ac{TTI} (potential collision within 1 second), with semantic-specific scores.}
	\label{tab:accuracy}
\end{table}

\subsection{Evasion}\label{sec:experiments:evasion}

Where \Cref{sec:experiments:tti} examined the magnitude and spatial accuracy of the predicted inverse \ac{TTI} maps, a control policy such as described in \Cref{sec:methodology:control} acts upon directional cues.
The evasion direction is based on the estimated motion vector of obstacles relative to the agent.
With a fixed control policy, the accuracy of this estimate represents an approach's ability to avoid oncoming obstacles.
Over the test set, we evaluate the average error magnitude between the estimated and ground truth motion vectors, and report the values in \Cref{tab:angaccuracy}.
\begin{table}[h]
	\centering
	\footnotesize
	\setlength\tabcolsep{3pt}
	\begin{tabular}{lrrrr}
		\toprule
		Approach & AAE & AAE (top 10\%)\\
		\midrule
		Inverse depth (baseline)		& \ang{16.58} & \ang{27.54}\\
		EVDodge (with GT optical flow)	& \ang{25.80} & \ang{22.28}\\
		Ours							& \textbf{\ang{2.65}} & \textbf{\ang{4.93}}\\
		Ours (dynamic)					& \textit{\ang{10.70}} & \textit{\ang{17.11}}\\
		\bottomrule
	\end{tabular}
	\caption{Average angle error for estimated trajectory of dangerous obstacles.
	`Top 10\%' represents the values with the top 10\% motion vector magnitudes}
	\label{tab:angaccuracy}
\end{table}
The table shows that all our approach much more accurately estimates the direction with the most associated danger.
Those in the top 10\% of ground truth magnitudes are particularly important as this relates to having the least time to avoid a collision.
This accuracy is possible through being able to filter out objects which may be in close proximity but are not currently dangerous.

\addtolength{\textheight}{-0.6cm}

\section{CONCLUSIONS}\label{sec:conclusion}

In this paper, we proposed that the fusion of events and depth data enable better collision avoidance strategies than using a single modality.
Depth sensors, while accurate for simple static scenes, struggle with fast-moving objects.
On the other hand, event cameras have excellent temporal resolution and dynamic range, but are not effective with textureless surfaces.
Each sensor modality complements the failure cases of the other.
We presented an approach which overcomes these issues, and directly estimates \acf{TTI} for every pixel.
Our method outperforms those based on foreground/background segmentation, and we demonstrated that the resulting \ac{TTI} maps can subsequently be used to direct a control policy to circumvent collsions.
We believe that this is an important step in utilising the benefits of bio-inspired sensors.
To complement our approach, we provide an extensive supporting dataset with a wide variety of data types to support further research.

\printbibliography
\end{document}